\documentclass[runningheads, a4paper]{llncs}
\usepackage[breaklinks=true,letterpaper=true,colorlinks,bookmarks=false]{hyperref}
\usepackage{amssymb,epsfig,graphicx}
\usepackage{color}

\usepackage{amsmath}
\usepackage{epstopdf}
\usepackage{bbm}
\usepackage{wrapfig}
\usepackage{algpseudocode}
\usepackage{algorithm}
\usepackage{multirow}
\usepackage{bm}
\usepackage{tabularx} 
\usepackage{dsfont}

\usepackage{tabularx} 
\newcolumntype{Y}{>{\centering\arraybackslash}X}

\usepackage{array,booktabs,arydshln,xcolor} 

\usepackage{epstopdf}
\usepackage{algorithm}
\usepackage{multirow}
\usepackage[algo2e]{algorithm2e}
\usepackage[outercaption]{sidecap}    
\usepackage{float}
\usepackage{wrapfig}
\usepackage{subcaption}
\usepackage{capt-of}
\usepackage{booktabs}
\usepackage{varwidth}
\begin{document}

\mainmatter  

\title{TandemNet: Distilling Knowledge from Medical Images Using  Diagnostic Reports \\ as Optional Semantic References}

\titlerunning{TandemNet}
\author{Zizhao Zhang, Pingjun Chen, Manish Sapkota, Lin Yang}
\authorrunning{Zizhao Zhang, \emph{et al.}}
\institute{University of Florida}

\toctitle{Lecture Notes in Computer Science}
\tocauthor{Authors' Instructions}
\maketitle

\vspace{-.7cm}
\begin{abstract}
In this paper, we introduce the semantic knowledge of medical images from their diagnostic reports to provide an inspirational network training and an interpretable prediction mechanism with our proposed novel multimodal neural network, namely TandemNet. Inside TandemNet, a language model is used to represent report text, which cooperates with the image model in a tandem scheme. We propose a novel dual-attention model that facilitates high-level interactions between visual and semantic information and effectively distills useful features for prediction.
In the testing stage, TandemNet can make accurate image prediction with an optional report text input. It also interprets its prediction by producing attention on the image and text informative feature pieces, and further generating diagnostic report paragraphs. 
Based on a pathological bladder cancer images and their diagnostic reports (BCIDR) dataset, sufficient experiments demonstrate that our method effectively learns and integrates knowledge from multimodalities and obtains significantly improved performance than comparing baselines. 
\end{abstract}

\section{Introduction}
\vspace{-.2cm}
In medical image understanding, convolutional neural networks (CNNs) gradually become the paradigm for various problems \cite{greenspan2016guest}. Training CNNs to diagnose medical images primarily follows pure engineering trends in an end-to-end fashion. However, the principles of CNNs during training and testing is difficult to interpret and justify.
In clinical practice, domain experts teach learners by explaining findings and observations to make a disease decision rather than leaving learners to find clues from images themselves. 

Inspired by this fact, in this paper, we explore the usage of semantic knowledge of medical images from their diagnostic reports to provide explanatory supports for CNN-based image understanding. The proposed network learns to provide interpretable diagnostic predictions in the form of attention and natural language descriptions.
The diagnostic report is a common type of medical record in clinics, which is comprised of semantic descriptions about the observations of biological features.
Recently, we have witnessed rapid development in multimodal deep learning research  \cite{vinyals2015show,xu2016multimodal}. 
We believe the joint study of multimodal data is essential towards intelligent computer-aided diagnosis. However, only a dearth of related work exists \cite{shin2016learning,zhang2017mdnet}.

\begin{figure}[t]	
	\begin{center}
		\includegraphics[width=0.999\textwidth]{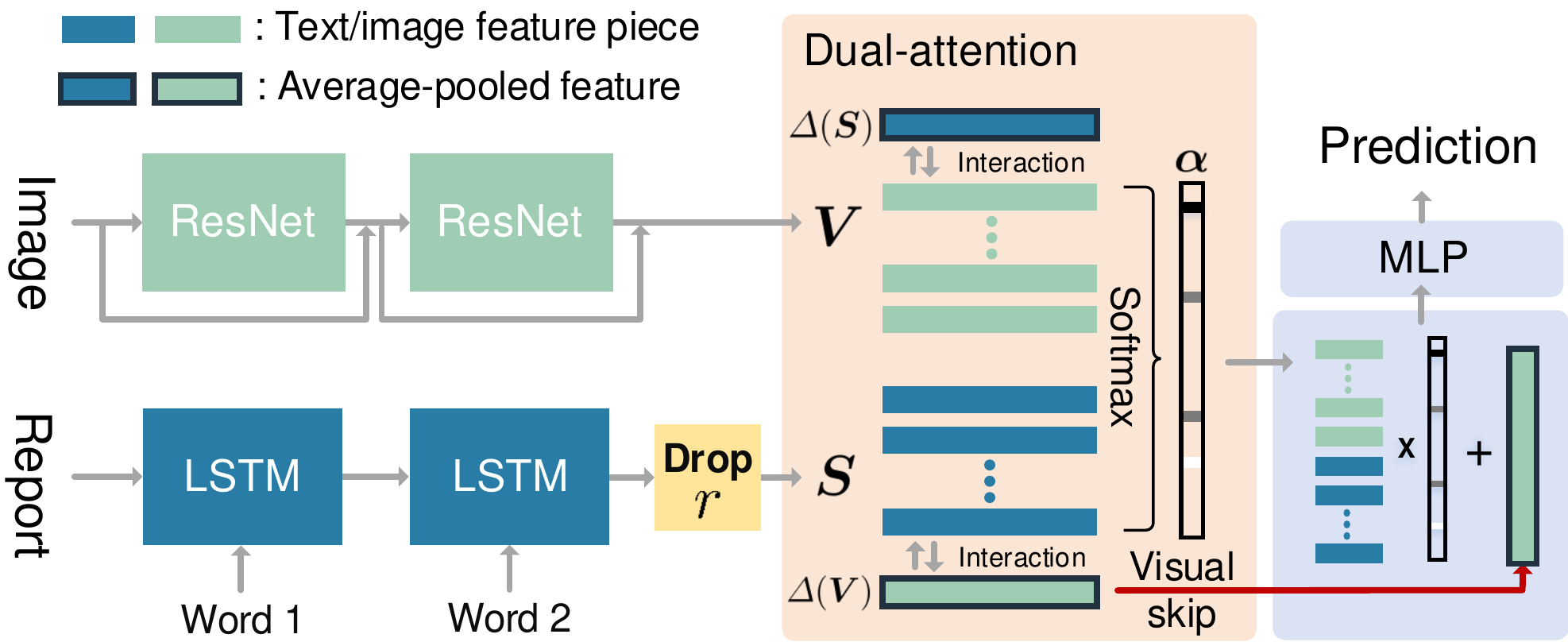}
	\end{center}
	\vspace{-.4cm}
	\caption{The illustration of the TandemNet.} \label{fig:outline} 
	\vspace{-.6cm}
\end{figure}

To take advantage of the language modality, we propose a multimodal network that jointly learns from medical images and their diagnostic reports. Semantic information is interacted with visual information to improve the image understanding ability by teaching the network to distill informative features. We propose a novel dual-attention model to facilitate such high-level interaction. The training stage uses both images and texts. In the testing stage, our network can take an image and provide accurate prediction with an optional (i.e. with or without) text input. Therefore, the language and image models inside our network cooperate with one another in a tandem scheme to either single(images)- or double(image-text)-drive the prediction process. We refer to our proposed network as TandemNet.  Figure \ref{fig:outline} illustrates the overall framework. 

To validate our method, we cooperate with a pathologist to collect the BCIDR dataset. Sufficient experimental studies on BCIDR demonstrate the advantages of TandemNet.  Furthermore, by coupling visual features with the language model and fine-tuning the network using backpropagation through time (BPTT), TandemNet learns to automatically generate diagnostic reports. 
The rich outputs (i.e. attention and reports) of TandemNet have valuable meanings: providing explanations and  justifications for its diagnostic prediction and making this process interpretable to pathologists.

\vspace{-.4cm}
\section{Method}

\vspace{-.4cm}
\noindent
\textbf{CNN for image modeling} We adopt the (new pre-activated) residual network (ResNet) \cite{he2016identity} as our image model. The identity mapping in ResNet significantly improves the network generalization ability. There are many architecture variants of ResNet. 
We adopt the wide ResNet (WRN) \cite{zagoruyko2016wide} which has shown better performance and higher efficiency with much less layers. It also offers scalability of the network (number of parameters) by adjusting a widen factor (i.e. the channel of feature maps) and depth. We extract the output of the layer before average pooling as our image representation, denoted as $\bm V \in \mathbb{R}^{C{\times}G}$. The input image size is $224{\times}224$, so $G = 14{\times}14$. $C$ depends on the widen factor.


\noindent
\textbf{LSTM for language modeling} We adopt Long Short-Term Memory (LSTM) \cite{hochreiter1997long}  to model diagnostic report sentences. LSTM improves vanilla recurrent neural networks (RNNs) for natural language processing and is also widely-used for multimodal applications such as image captioning \cite{karpathy2015deep,vinyals2015show}.
It has a sophisticated unit design, which enables long-term dependency and greatly reduces the gradient vanishing problem in RNNs \cite{pascanu2013difficulty}. Given a sequence of words $\{\bm x_1,...,\bm x_n\}$,
LSTM  reads the words one at a time and maintains a memory state $\bm m_t \in \mathbb{R}^{D}$ and a hidden state $\bm h_t \in \mathbb{R}^{D}$. At each time step, LSTM updates them by 
\vspace{-.1cm}
\begin{equation}
\bm h_t, \, \bm m_t = \text{LSTM}(\bm x_t, \bm h_{t-1}, \bm m_{t-1}), \vspace{-.1cm}
\end{equation} 
where $\bm x_t \in \mathbb{R}^{K}$ is an input word, which is computed by firstly encoding it as a one-hot vector and then multiplied by a learned word embedding matrix.  

The hidden state is a vector encoding of sentences. The treatment of it varies from problems. For example, in image captioning, a multilayer perceptron (MLP) is used to decode it as a predicted word at each time step. In machine translation \cite{luong2015effective}, all hidden states could be used. 
A medical report is more formal than a natural image caption. It usually describes multiple types of biological features structured by a series of sentences. It is important to represent all feature descriptions but maintain the variety and independence among them. To this end, we extract the hidden state of every feature description (in our implementation, it is achieved by adding a special token at the end of each sentence beforehand and extracting the hidden states at all the placed tokens). In this way, we obtain a text representation matrix $\bm S = [\bm h_1,...,\bm h_N] \in \mathbb{R}^{D{\times}N}$ for $N$ types of feature descriptions. This strategy has more advantages: it enables the network to adaptively select useful semantic features and determine respective feature importance to disease labels (as shown in experiments).


\noindent
\textbf{Dual-attention model} The attention mechanism \cite{xu2015show,luong2015effective} is an active topic in both computer vision and natural language communities. Briefly, it gives networks the ability to generate attention on parts of the inputs (like visual attention in the brain cortex), which is achieved by computing a context vector with attended information preserved.

Different from most existing approaches that study attention on images or text, given the image representation $\bm V$ and the report representation $\bm S$\footnote{The two matrices are firstly embedded through a $1{\times}1$ convolutional layer with Tanh.}, our dual-attention model can generate attention on important image regions and sentence parts simultaneously. Specifically, we define the attention function $f_{att}$ to compute a piece-wise weight vector $\bm \alpha$ as
\begin{equation}
\bm e = f_{att} (\bm V, \bm S), \;\;\; \bm \alpha_i = \frac{\text{exp}(\bm e_i)}{\sum_{i} \text{exp}(\bm e_i)},
\end{equation} 
where $\bm \alpha  \in \mathbb{R}^{G+N}$ has individual weights for visual and semantic features (i.e. $\bm V$ and $\bm S$). $f_{att}$ is specifically defined as follows:
\begin{equation}
\begin{split}
\bm z_{s\rightarrow v}  &=  \tanh (\bm W_v  \bm V + (\bm W_{s'} \Delta(\bm S)) \mathds{1}_v^T ), \\
\bm z_{v \rightarrow s}  &=  \tanh (\bm W_s  \bm S + (\bm W_{v'} \Delta(\bm V)) \mathds{1}_s^T), \\
\bm e &= \bm w^T [\bm z_{s\rightarrow v} ; \bm z_{v \rightarrow s}] + \bm b,
\end{split}
\end{equation} 
where $\bm W_v,  \bm W_{v'} \in \mathbb{R}^{M {\times} C}$ and $\bm W_s,  \bm W_{s'} \in \mathbb{R}^{M {\times} D}$ are parameters to be learned to compute $\bm z_{s\rightarrow v} \in \mathbb{R}^{M{\times}G} \text{ and } z_{v \rightarrow s} \in \mathbb{R}^{M{\times}N}$, and $\bm w, \bm b \in  \mathbb{R}^{M}$. $\mathds{1}_v \in  \mathbb{R}^{G}$ and $\mathds{1}_s \in  \mathbb{R}^{N}$  are vectors with all elements to be one. $\Delta$ denotes the global average-pooling operator on the last dimension of $\bm V$ and $\bm S$. $[\,; ]$ denotes the concatenation operator.
Finally, we obtain a context vector $\bm c \in \mathbb{R}^{M}$ by
\vspace{-.2cm}
\begin{equation}
\bm c = \bm O \, \bm \alpha = \sum_{i=1}^{G}  \alpha_i \bm V_ i+ \sum_{j=G+1}^{G+N}  \alpha_j \bm S_j  , \text{ where } \bm {O} = [\bm V; \bm S]. \vspace{-.2cm}
\end{equation}
In our formulation, the computation of image and text attention is mutually dependent and conducts high-level interactions. The image attention is conditioned on the global text vector $\Delta(\bm S)$ and the text attention is conditioned on the global image vector $\Delta(\bm V)$. When computing the weight vector $\bm \alpha$, both information contributes through $\bm z_{s\rightarrow v} \text{ and } \bm z_{v \rightarrow s}$. We also consider extra configurations: computing two $\bm e$ by two $\bm w$, and then concatenate them to compute $\bm \alpha$ with one softmax or compute two $\bm \alpha$ with two softmax functions. Both configurations underperform ours. We conclude that our configuration is optimal for the visual and semantic information to interact with each other.

Intuitively, our dual-attention mechanism encourages better alignment of visual information with semantic information piecewise, which thereby improves the ability of TandemNet to discriminate useful features for attention computation. We will validate this experimentally.

\noindent
\textbf{Prediction module} To improve the model generalization, we propose two effective techniques for the prediction module of the dual-attention model.

\noindent
1) \textit{Visual skip-connection} The probability of a disease label $p$ is computed as 
\vspace{-.2cm}
\begin{equation}
\label{eq:mlp}
p = \text{MLP}(\bm c +  \Delta(\bm V)). \vspace{-.2cm}
\end{equation}
The image feature $\Delta(\bm V)$ skips the dual-attention model and is directly added onto $\bm c$ (see Figure \ref{fig:outline}).  During backpropagation, this skip-connection directly passes gradients for the loss layer to the CNN, which prevents possible gradient vanishing in the dual-attention model from obstructing CNN training.

\noindent
2) \textit{Stochastic modality adaptation} We propose to stochastically ``abandon'' text information during training. This strategy generalizes TandemNet to make accurate prediction with absent text. 
Our proposed strategy is inspired by Dropout and the stochastic depth network \cite{huang2016deep}, which are effective for model generalization. 
Specifically,  we define a drop rate $r$ as the probability to remove (zero-out) the text part $\bm S$  during the entire network training stage. Thus, based to the principle of Dropout, $\bm S$ will be scaled by $1-r$ if text is given in testing. 

The effects of these two techniques are discussed in experiments.

\begin{table*}[t] 
	\begin{minipage}[b]{0.20\linewidth}
			\centering
		\begin{tabularx}{.99\textwidth}{c|c|c}
			
			\cline{1-3} 
			\multirow{2}{*}{Method} & \multicolumn{2}{c}{Accuracy ($\%$)} \\ \cline{2-3}
			& w/o text &       w/ text        \\ \cline{1-3}			
			WRN16-4                                   &    75.4    &           -            \\ 
			ResNet18-TL                               &    79.4    &           -            \\ \cline{1-3}
			TandemNet-WVS						&	79.4		&	85.6	\\
			TandemNet                                 &   82.4    &     89.9		\\
			TandemNet-TL                              &   84.9     &     88.6     		\\ 
			\cline{1-3}
		\end{tabularx}
		\vspace{.0cm}
		\captionsetup{width=2.1\linewidth}
		\captionof{table}{The quantitative evaluation (averaged on 3 trials). The first block shows standard CNNs so text is  irrevelent.\label{tab:comp}}
	\end{minipage}\hfill
	\begin{minipage}[b]{0.57\linewidth}
		\centering
		\includegraphics[width=0.90\textwidth,height=0.47\textwidth]{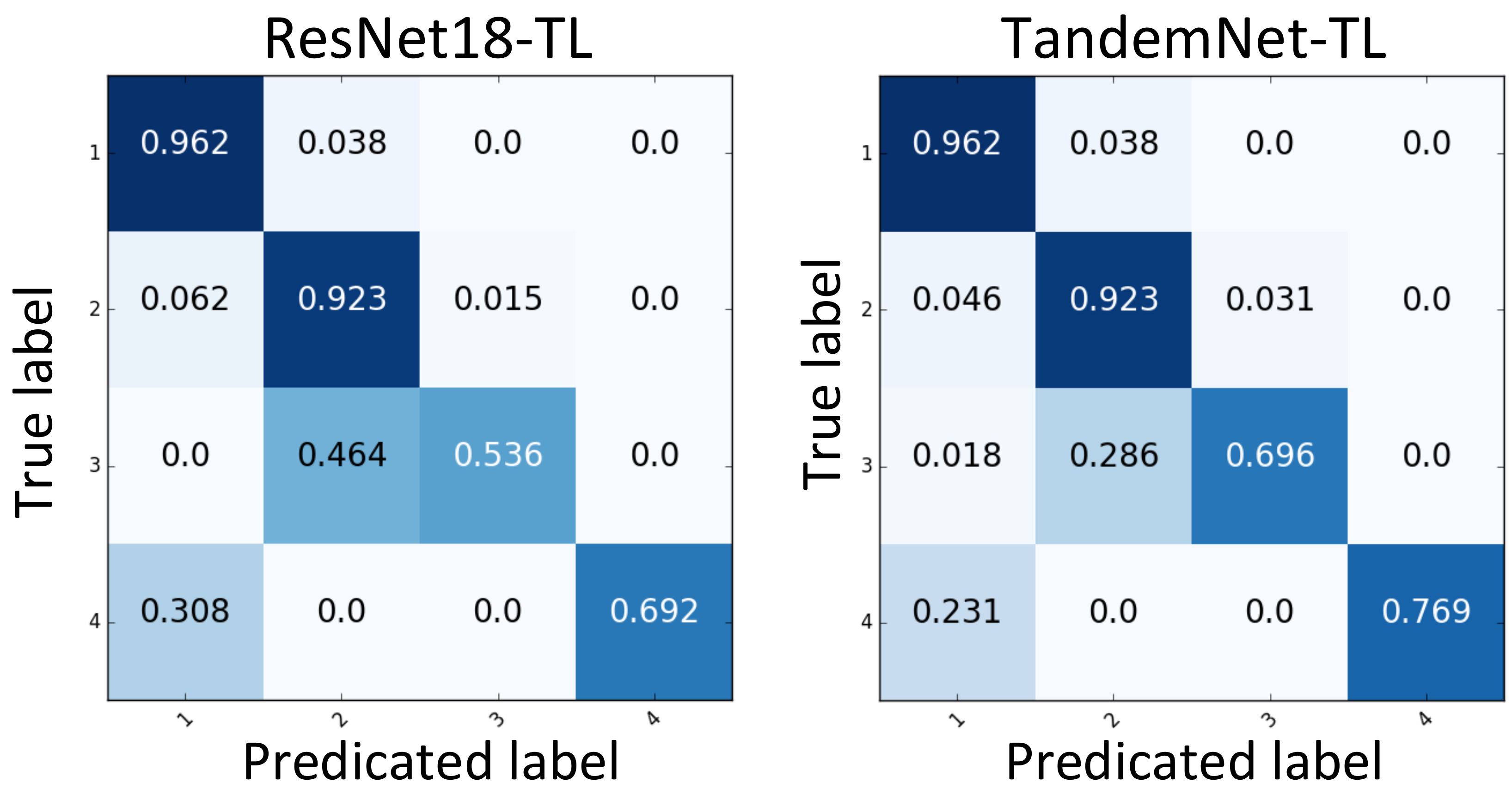} 
			\vspace{.2cm}
		\captionsetup{width=0.9\linewidth}
		\captionof{figure}{The confusion matrices of two compared methods ResNet18-TL and TandemNet-TL (w/o text) in Table \ref{tab:comp}. \label{fig:conf}}
		
		\label{fig:image}
	\end{minipage}
	\vspace{-.9cm}
\end{table*}

\vspace{-.4cm}
\section{Experiments}
\vspace{-.4cm}
\noindent
\textbf{Dataset} To collect the BCIDR dataset, whole-slide images were taken using a 20X objective from hematoxylin and eosin (H$\&$E) stained sections of bladder tissue extracted from a cohort of 32 patients at risk of a papillary urothelial neoplasm. From these slides, 1,000 $500{\times}500$ RGB images were extracted randomly close to urothelial regions (each patient's slide yields a slightly different number of images). For each of these images, the pathologist then provided a paragraph describing the disease state. Each paragraph addresses five types of cell appearance features, namely the state of \textit{nuclear pleomorphism}, \textit{cell crowding}, \textit{cell polarity}, \textit{mitosis}, and \textit{prominence of nucleoli} (thus $N{=}5$). Then a conclusion is decided for each image-text pair, which is comprised of four classes, i.e. \textit{normal} tissue, \textit{low-grade} (papillary urothelial neoplasm of low malignant potential) carcinoma, \textit{high-grade} carcinoma, and \textit{insufficient information}. Following the same procedure, four doctors (not experts in the bladder cancer) wrote additional four descriptions for each image. They also refer to the pathologist's description to make sure their annotation accuracy. Thus there are five ground-truth reports per image and $5,000$ image-text pairs in total. Each report varies in length between 30 and 59 words. We randomly split $20\%$ (6/32) of patients including $1,000$ samples as the testing set and the remaining $80\%$ of patients including $4,000$ samples ($20\%$ as the validation set for model selection) for training. We subtract the data RGB mean and augment through clip, mirror and rotation.

\noindent
\textbf{Implementation details} Our implementation is based on Torch7. We use a small WRN with $\text{depth}{=}16$ and $\text{widen-factor}{=}4$ (denoted as WRN16-4), resulting in $2.7$M parameters and $C{=}256$. We use dropout with $0.3$ after each convolution. We use $D{=}256$ for LSTM, $M{=}256$, and $K{=}128$. We use SGD with a learning rate $1e{-}2$ for the CNN (used likewise for standard CNN training for comparison) and Adam with $1e{-}4$ for the dual-attention model, which are multiplied by $0.9$ per epoch.  We also limit the gradient magnitude of the dual-attention model to $0.1$ by normalization \cite{pascanu2013difficulty}. 

\noindent
\textbf{Diagnostic prediction evaluation}
Table \ref{tab:comp} and Figure \ref{fig:conf} show the quantitative evaluation of TandemNet. For comparison with CNNs, we train a WRN16-4 and also a ResNet18 (has 11M parameters) pre-trained on ImageNet\footnote{Provided by \url{https://github.com/facebook/fb.resnet.torch}}. We found transfer learning is beneficial. To test this effect in TandemNet, we replace WRN16-4 with a pre-trained ResNet18 (TandemNet-TL). As can be observed, TandemNet and TandemNet-TL significantly improve WRN16-4 and ResNet18-TL when only images are provided. We observe TandemNet-TL slightly underperforms TandemNet when text is provided with multiple trails. We hypothesize that it is because fine-tuning a model pre-trained on a complete different natural image domain is relatively hard to get aligned with medical reports in the dual-attention model. From Figure \ref{fig:conf}, \textit{high grade} (label id 3) is more likely to be misclassified as \textit{low grade} (2) and some \textit{insufficient information} (4) is confused with \textit{normal} (1).

\begin{figure}[t]	
	\begin{center}
		
		\begin{subfigure}[b]{0.49\textwidth}
			\centering
			\includegraphics[width=0.999\textwidth,height=0.53\textwidth]{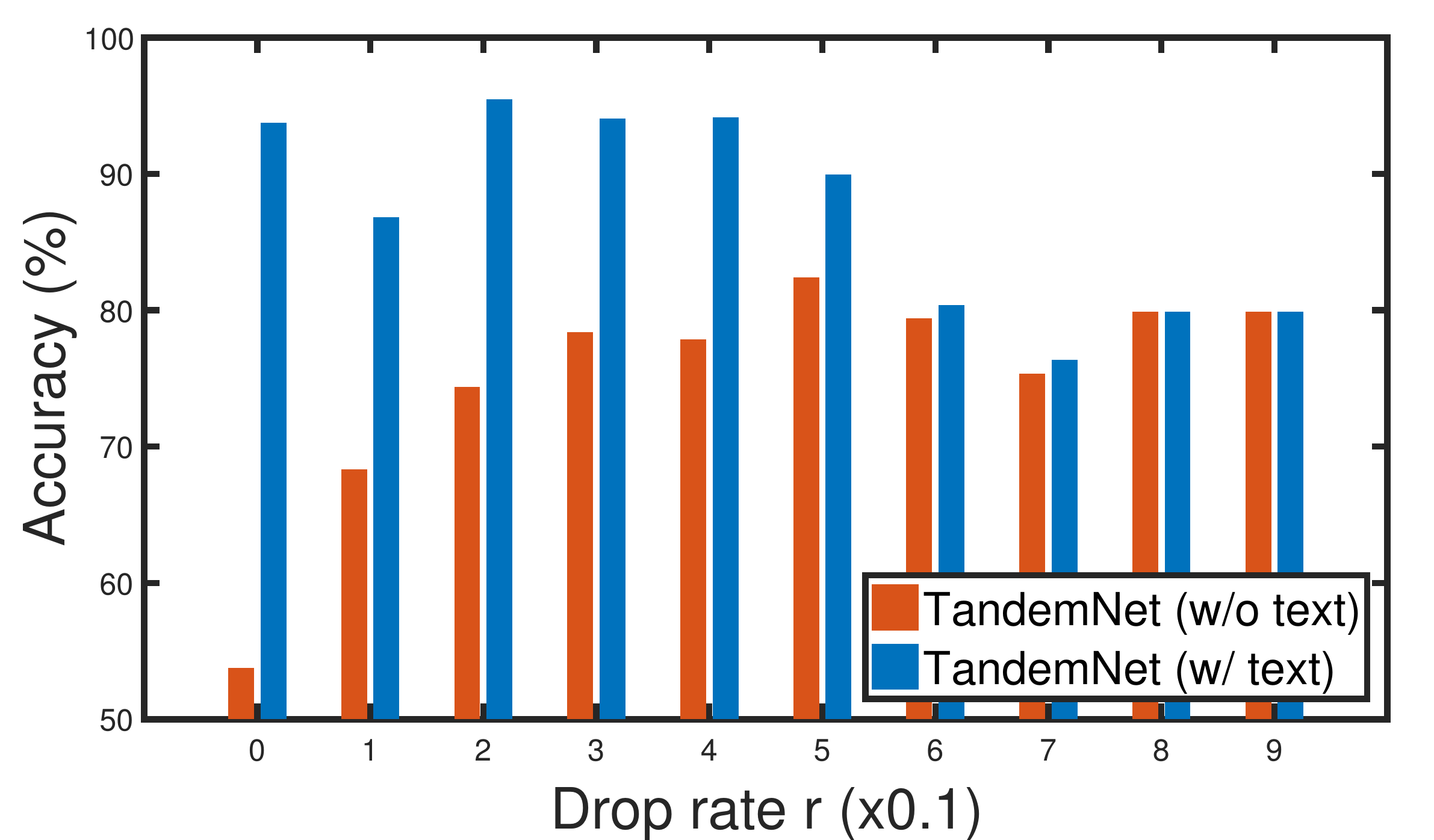}
		\end{subfigure}
		\begin{subfigure}[b]{0.49\textwidth}
			\centering
			\includegraphics[width=0.999\textwidth,height=0.53\textwidth]{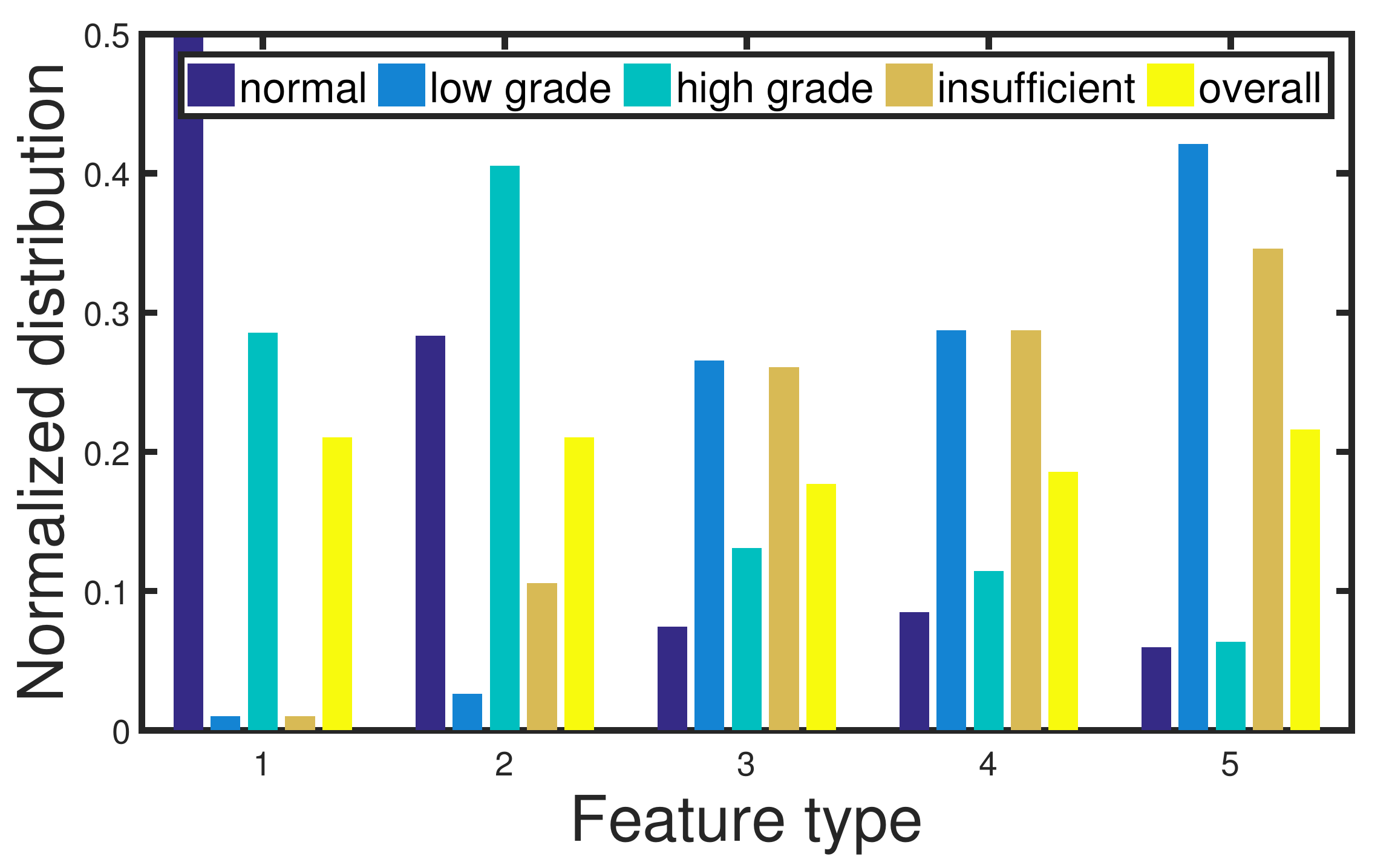}
		\end{subfigure}
	\end{center}
	\vspace{-.7cm}
	\caption{Left: The accuracy with varying drop rates.  Right: The averaged text attention per feature type (and overall) to each disease label. The feature type is specified in the text of dataset introduction (in order). } \label{fig:text_attention} 	\vspace{-.3cm}
\end{figure}

\begin{figure}[t]	
	\begin{center}
		\includegraphics[width=0.99\textwidth,height=0.3\textwidth]{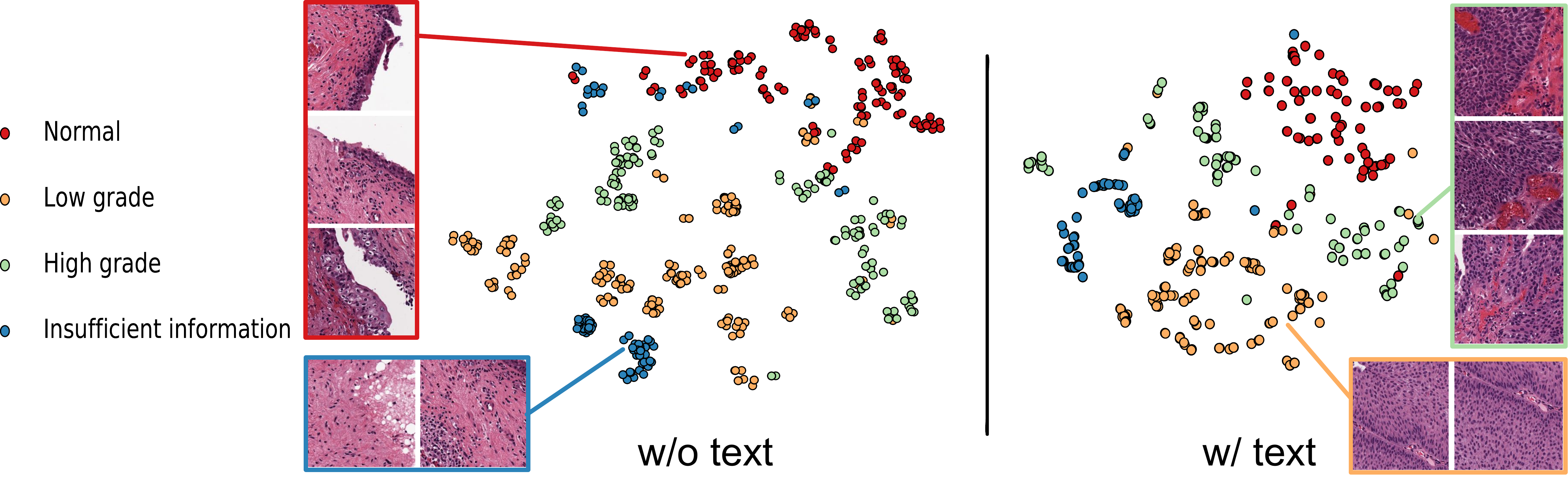}
	\end{center}
	\vspace{-.6cm}
	\caption{The t-SNE visualization of the MLP input. Each point is a test sample. The embeddings with text (right) results in better distribution. } \label{fig:tsne}
	\vspace{-.6cm}
\end{figure}

We analyze the text drop rate in Figure \ref{fig:text_attention} (left). When the drop rate is low, the model obsessively uses text information, so it achieves low accuracy without text. 
When the drop rate is high, the text can not be well adapted, resulting in decreased accuracy with or without text. The drop rate of $0.5$ performs best and thereby is used in this paper.  As illustrated in Figure \ref{fig:text_attention}, we found that the classification of text is easier than images, therefore its accuracy is much higher. However, please note that the primary aim of this paper is to use text information only at the training stage. While at the testing stage, the goal is to accurately classify images without text.

In Eq. (\ref{eq:mlp}), one question that may arise is that, when testing without text, whether it is merely $\Delta(\bm V)$ from the CNN that produces useful features rather than $\bm c$ from the dual-attention model (since the removal (zero-out) of $\bm S$ could possibly destroy the attention ability). To validate the actual role of $\bm c$,  we remove the visual skip-connection and train the model (denoted as TandemNet-WVS in Table \ref{tab:comp}) and it improves ResNet16-4 by $4\%$ without text. 
The qualitative evaluation below also validates the effectiveness of the dual-attention model. Additionally, we use the (t-distributed Stochastic Neighbor Embedding) t-SNE dimensionality reduction technique to examine the input of MLP in Figure \ref{fig:tsne}.

\begin{figure}[t]	
	\begin{center}
		\includegraphics[width=0.99\textwidth]{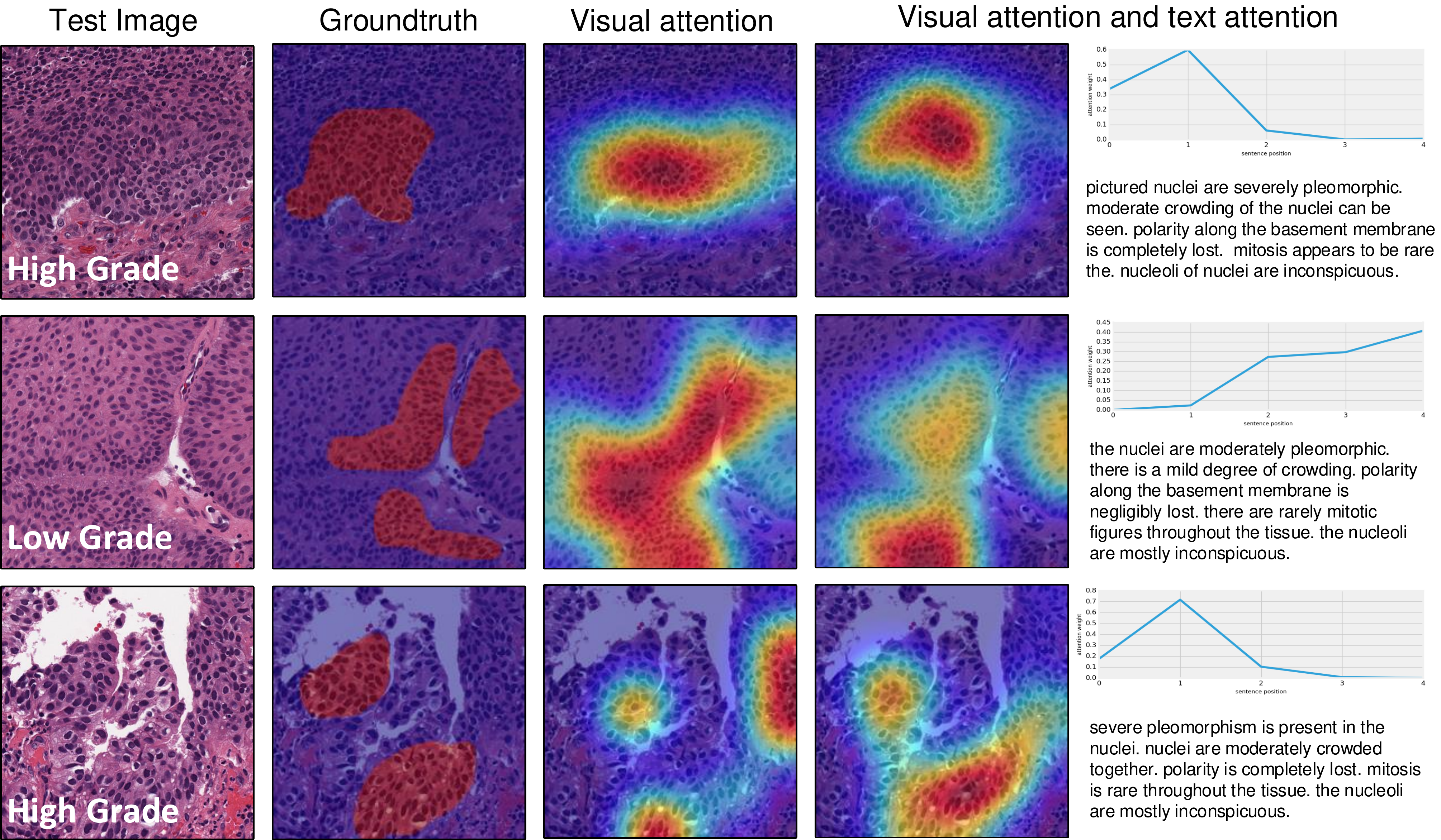}
	\end{center}
	\vspace{-.5cm}
	\caption{From left to right: Test images (the bottom shows disease labels), pathologist's annotations, visual attention w/o text. visual attention and corresponding text attention (the bottom shows text inputs). Best viewed in color.} \label{fig:attention}
	\vspace{-0.5cm}
\end{figure}

\noindent
\textbf{Attention analysis}
We  visualize the attention weights to show how TandemNet captures image and text information to support its prediction (the image attention map is computed by upsampling the $G{=}14{\times}14$ weights of $\bm \alpha$ to the image space). 
To validate the visual attention, without notifying our results beforehand, we ask the pathologist to highlight regions of some test images they think are important. Figure \ref{fig:attention} illustrates the performance. Our attention maps show surprisingly high consistency with pathologist's annotations. The attention without text is also fairly promising, although it is less accurate than the results with text. Therefore, we can conclude that TandemNet effectively uses semantic information to improve visual attention and substantially maintains such attention capability though the semantic information is not provided. 
The text attention is shown in the last column of Figure \ref{fig:attention}. We can see that our text attention result is quite selective in only picking up useful semantic features.

Furthermore, the text attention statistics over the dataset provides particular insights into the pathologists' diagnosis. We can investigate which feature contributes the most to which disease label (see Figure \ref{fig:text_attention} (right)). For example, \textit{nuclear pleomorphism} (feature type 1) shows small effects on the \textit{low-grade} disease label. \textit{cell crowding} (2) has large effects on \textit{high-grade}. We can justify the reason of text attention by closely looking at images of Figure \ref{fig:attention}: \textit{high grade} images have obvious high \textit{cell crowding} degree. Moreover, this result strongly demonstrates the successful image-text alignment of our dual-attention model.

\begin{figure}[t]	
	\begin{center}
		\includegraphics[width=0.99\textwidth]{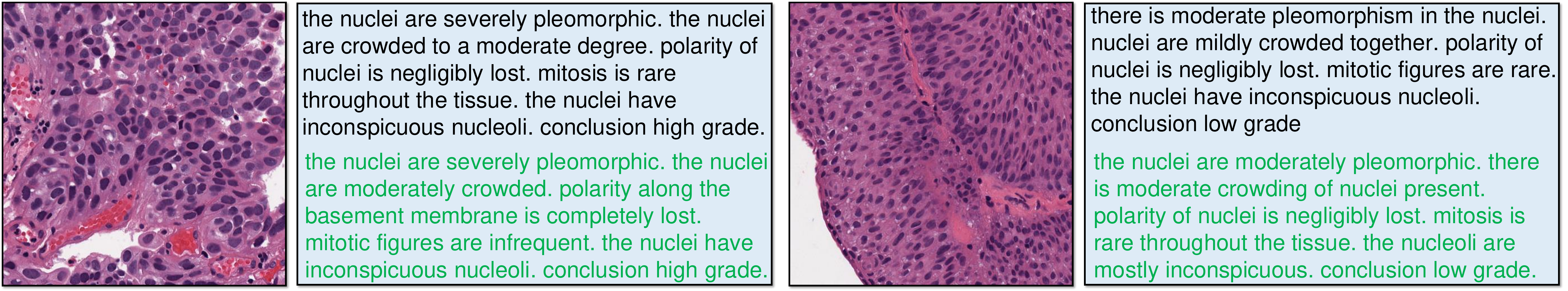}
	\end{center}
	\vspace{-.6cm}
	\caption{The pathologist's annotations are in black and the automatic results of TandemNet are in green, which accurately describe the semantic concepts.} \label{fig:report} 
	\vspace{-.6cm}
\end{figure}

\noindent
\textbf{Image report generation} We fine-tune TandemNet using BPTT as an extra supervision and use the visual feature $\Delta(\bm V)$ as the input of LSTM at the first time step\footnote{We freeze the CNN for the whole training and the dual-attention model for the first $5$ epochs, and then fine-tune with a smaller learning rate, $5e{-}5$. }. We direct readers to \cite{karpathy2015deep} about detailed LSTM training for image captioning. Figure \ref{fig:report} shows our promising results compared with pathologist's descriptions. We leave the full report generation task as a future study \cite{zhang2017mdnet}.

\vspace{-.4cm}
\section{Conclusion}
\vspace{-.3cm}
This paper proposes a novel multimodal network, TandemNet, which can jointly learn from medical images and diagnostic reports and predict in an interpretable scheme through a novel dual-attention mechanism.
Sufficient and comprehensive experiments on BCIDR demonstrate that TandemNet is favorable for more intelligent computer-aided medical image diagnosis.

\vspace{-.4cm}

\bibliographystyle{splncs}
\bibliography{reference}

\vspace{-.5cm}
\end{document}